\documentclass[10pt,twocolumn,letterpaper]{article}

\usepackage[pagenumbers]{cvpr}

\usepackage{graphicx}
\usepackage{amsmath}
\usepackage{amssymb}
\usepackage{booktabs}
\usepackage{enumitem}
\usepackage[accsupp]{axessibility}
\usepackage[pagebackref,breaklinks,colorlinks]{hyperref}

\usepackage[capitalize]{cleveref}
\crefname{section}{Sec.}{Secs.}
\Crefname{section}{Section}{Sections}
\Crefname{table}{Table}{Tables}
\crefname{table}{Tab.}{Tabs.}

\usepackage[dvipsnames]{xcolor}

\usepackage{amsmath}
\usepackage{pifont}
\usepackage{amssymb}
\newcommand{\A}{{\mathcal A}}
\newcommand{\Aprime}{{\mathcal A^{\prime}}}
\newcommand{\Sprime}{{\mathcal S^{\prime}}} 
\newcommand{\D}{{\mathcal D}}
 
\newcommand{\mc}[3]{\multicolumn{#1}{#2}{#3}}
\definecolor{Burgundy}{RGB}{144,0,32}
\newcommand{\red}[1]{\textcolor{Burgundy}{#1}}
\definecolor{ForestGreen}{RGB}{34,139,34}
\newcommand{\green}[1]{\textcolor{ForestGreen}{#1}}

\newcommand{\lcls}{{\mathcal{L}_{\text{cls}}}}
\newcommand{\lat}{{\mathcal{L}_{\text{attr}}}}
\newcommand{\lob}{{\mathcal{L}_{\text{obj}}}}
\newcommand{\lse}{{\mathcal{L}_{\text{seen}}}}
\newcommand{\lun}{{\mathcal{L}_{\text{unseen}}}}
\newcommand{\Iat}{{I_{\text{attr}}}}
\newcommand{\Iob}{{I_{\text{obj}}}}
\newcommand{\fat}{{f_{\text{attr}}}}
\newcommand{\fob}{{f_{\text{obj}}}}
\newcommand{\vao}{{v_\text{attr,obj}}}

\newcommand{\wao}{{w_\text{attr,obj}}}
\newcommand{\waop}{{w_\text{attr,obj}^{\prime}}}
\newcommand{\va}{{v_\text{attr}}}
\newcommand{\vap}{{v_\text{attr}^{\prime}}}
\newcommand{\vo}{{v_\text{obj}}}
\newcommand{\vop}{{v_\text{obj}^\prime}}
\newcommand{\wa}{{w_\text{attr}}}
\newcommand{\wo}{{w_\text{obj}}}

\newcommand{\yo}{{y_\text{obj}}}
\newcommand{\ma}{{m_\text{attr}}}
\newcommand{\mo}{{m_\text{obj}}}
\newcommand{\mop}{{m_\text{obj}^\prime}}
\newcommand{\map}{{m_\text{attr}^{\prime}}}

\begin{document}

\title{Disentangling Visual Embeddings for Attributes and Objects}

\author{Nirat Saini \quad\qquad Khoi Pham \quad\qquad Abhinav Shrivastava\\[0.5em]
University of Maryland, College Park
}
\maketitle

\begin{abstract}
  We study the problem of compositional zero-shot learning for object-attribute recognition.  Prior works use visual features extracted with a backbone network, pre-trained for object classification and thus do not capture the subtly distinct features associated with attributes. To overcome this challenge, these studies employ supervision from the linguistic space,  and use pre-trained  word  embeddings  to  better  separate  and  compose attribute-object pairs for recognition. Analogous to linguistic embedding space, which already has unique and agnostic embeddings for object and attribute, we shift the focus back to the visual space and propose a novel architecture that can disentangle attribute and object features in the visual space.  We use visual decomposed features to hallucinate embeddings that are representative for the seen and novel compositions to better regularize the learning of our model. Extensive  experiments  show  that  our  method outperforms existing work with significant margin on three datasets:  MIT-States,  UT-Zappos,  and  a  new  benchmark created based on VAW. The code, models, and dataset splits are publicly available at~\url{https://github.com/nirat1606/OADis}.
\end{abstract}

\vspace{-1em}
\section{Introduction}
\label{sec:intro}
Objects in the real world can appear with different properties, \ie, different color, shape, material, etc. For instance, an apple can be red or green, cut or peeled, raw or ripe, and even dirty or clean. Understanding object properties can greatly benefit various applications, \eg, robust object detection~\cite{cobe,obj,ob1,ob3}, human object interaction~\cite{hoi, hoi1, hoi2}, and activity recognition~\cite{object_centric_new,perceptual_causality,epic_states1,epic_states,joint_dis, fathi}. Since the total number of possible attribute-object pairs in the real world is prohibitively large, it is impractical to collect image examples and train multiple classifiers. Prior works proposed compositional learning, \ie, learning to compose knowledge of known attributes and object concepts to recognize a new attribute-object composition. Datasets such as MIT-States~\cite{mitstates} and UT-Zappos~\cite{utzappos} are commonly used to study this task, with joint attribute-object recognition for a diverse, yet limited set of objects and attributes.

Compositional learning refers to combining simple primitive concepts to understand a complex concept.
\begin{figure}
\centering
\vspace{-.1in}
\includegraphics[width=\linewidth]{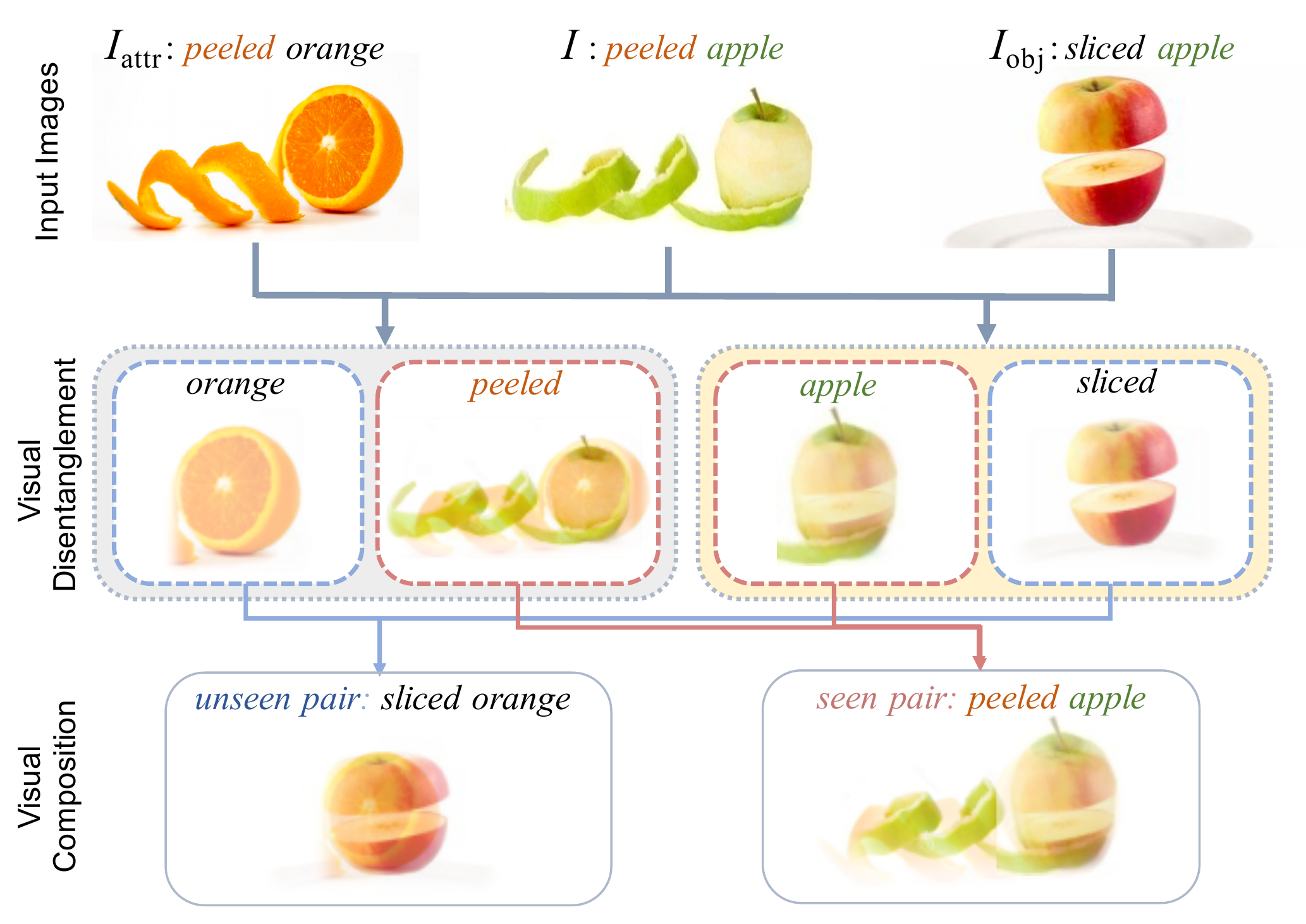}
\vspace{-.2in}
\caption{\textbf{Method illustration}: Given an input image $I$ of \texttt{peeled} \texttt{apple}, we use two other images: (1) one with same object, different attribute $\Iob$ - \texttt{sliced} \texttt{apple}, (2) one with same attribute, different object $\Iat$ - \texttt{peeled} \texttt{orange}. We propose a novel architecture that takes $I$ and $\Iat$, and extracts their visual similarity features for \texttt{peeled} and visual dissimilarity features for \texttt{orange}. Similarly, using $I$ and $\Iob$, the visual similarity features for \texttt{apple}, and the dissimilarity features for \texttt{sliced} can be extracted. We compose these primitive visual features to hallucinate a seen pair \texttt{peeled} \texttt{apple}, and a novel unseen pair \texttt{sliced} \texttt{orange} to be used for regularizing our embedding space. Note that this is a visualization of embedding space composition, we do not generate images.}
\label{fig:teaser_v1}
\vspace{-0.2in}
\end{figure}
 This idea dates back to Recognition and Composition theory by Biederman~\cite{bied}, and early work in the visual domain by Hoffman~\cite{hoff}, which proposed recognition by parts for pose estimation.
Prior works explore compositionality to a certain degree, \eg, via feature sharing and shared embeddings space. Among them, most works use linguistically inspired losses to separate attributes and objects in the shared embedding space, then use that primitive knowledge to compose new complex pairs. Using linguistic embeddings is helpful since: (1) there is a clear distinction between attribute and object in the embedding space, and (2) these embeddings already contain semantic knowledge of similar objects and attributes, which is helpful for composition.
However, unlike word embedding, it is difficult to discriminate the object and attribute in the visual embedding space. 

This is due to the fact that image feature extractor is usually pre-trained for object classification, often along with image augmentation (\eg, color jitter) that tends to produce attribute-invariant image representation, thus does not learn objects and attributes separately.
In this paper, we propose a new direction that focuses on \underline{\textit{visual cues}}, instead of using linguistic cues explicitly for novel compositions. 

Analogous to linguistic embedding, our work focuses on disentangling attribute and object in the visual space. Our method, Object Attribute Disentanglement (OADis), learns distinct and independent visual embeddings for \texttt{peeled} and \texttt{apple} from the visual feature of \texttt{peeled} \texttt{apple}. As shown in Figure~\ref{fig:teaser_v1}, for image $I$ of \texttt{peeled} \texttt{apple}, we use two other images: one with same object and different attribute $\Iob$ (\eg, \texttt{sliced} \texttt{apple}), and one with same attribute and different object $\Iat$ (\eg, \texttt{peeled} \texttt{orange}). OADis takes $I$ and $\Iob$ and learns the similarity (\texttt{apple}) and dissimilarity (\texttt{sliced}) of the second image with respect to the first one. Similarly, using $I$ and $\Iat$, the commonality between them (\texttt{peeled}) and the left out dissimilarity (\texttt{orange}) can also be extracted. Further, composition of these extracted visual primitives are used to hallucinate seen and unseen pair, \texttt{peeled} \texttt{apple} and \texttt{sliced} \texttt{orange} respectively. 

\vspace{-0.3pt}
\begin{figure*}
\vspace{-0.1in}
\centering
\includegraphics[width=0.95\linewidth]{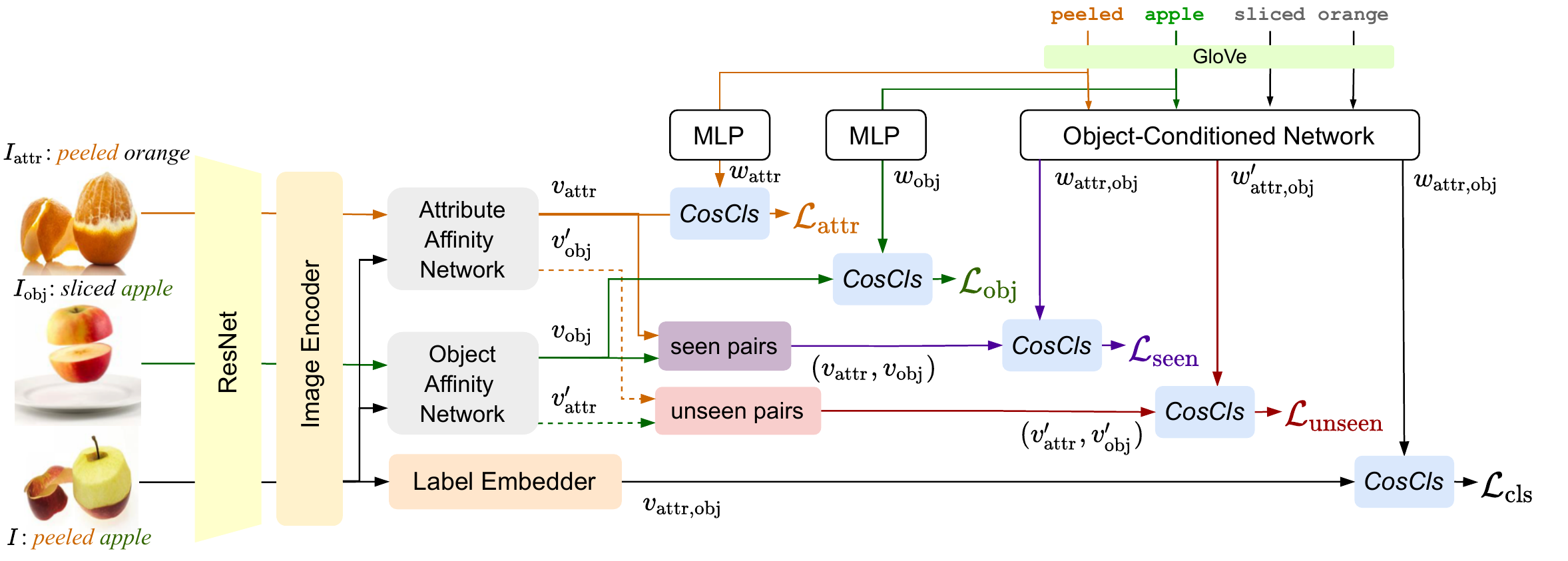}
\vspace{-.1in}
\caption{\textbf{System Overview}: Given an image $I$, for \texttt{peeled} \texttt{apple}, we consider two images:, one with same object: $\Iob$, \texttt{sliced} \texttt{apple}, and one with same attribute, $\Iat$ \texttt{peeled} \texttt{orange}. (1) The Object-Conditioned Network composes pair word embedding, using GloVe word embeddings for labels. (2) Label Embedder uses the image $I$ and embeds visual feature $\vao$ along with word embedding $\wao$, using loss $\lcls$. (3) Attribute Affinity Network and Object Affinity Network, disentangles the same attribute and object from the pair of images $I, \Iat$ and  $I , \Iob$ respectively. Disentangled visual features for \texttt{peeled} ($\va$) and \texttt{apple} ($\vo$) are used along with word embeddings of attribute ($\wa$) and objects ($\wo$), to compute $\lat$ and $\lob$. (4) Using disentangled features, we compose seen pair \texttt{peeled} \texttt{apple} ($\va,\vo$) and unseen pair \texttt{sliced} \texttt{orange} ($\vap,\vop$), for composition losses $\lse$ and $\lun$.} 
\label{fig:system}
\vspace{-.2in}
\end{figure*}

For compositional learning, it is necessary to decompose first before composing new unseen attribute-object pairs. As humans, we have the ability to imagine an unseen complex concept using previous knowledge of its primitive concepts. For example, if someone has seen a \texttt{clown} and a \texttt{unicycle}, they can imagine \texttt{clown} \texttt{on} \texttt{a} \texttt{unicycle} even if they have never seen this combination in real life~\cite{human_imagine, human_imagine2}. This quality of \textit{imagination} is the basis of various works such as GANs~\cite{stargan}, CLIP~\cite{clip} and DALL-E~\cite{dalle}. However, these works rely on larger datasets and high computation power for training. We study this idea of \textit{imagination} for a smaller setup by composing newer complex concepts using disentangled attributes and object visual features. Our work focuses on answering the question, \textit{can there be visual embedding of \texttt{peeled} and \texttt{apple}, disentangled separately from visual feature of \texttt{peeled} \texttt{apple}?} Our contributions are as follows:
\begin{itemize}[noitemsep,left=0pt]
    \item We propose a novel approach, OADis, to disentangle attribute and object visual features, where visual embedding for \texttt{peeled} is distinct and independent of embedding for \texttt{apple}.
    \item We compose unseen pairs in the visual space using the disentangled features. Following Compositional Zero-shot Learning (CZSL) setup, we show competitive improvement over prior works on standard datasets~\cite{mitstates,utzappos}.
    \item We propose a new large-scale benchmark for CZSL using an existing attribute dataset VAW~\cite{khoi}, and show that OADis outperforms existing baselines.
\end{itemize}

\vspace{-1em}
\section{Related Work}
\vspace{-0.03in}
\noindent \textbf{Visual Attributes.}
Visual attributes have been studied widely to understand visual properties and low-level semantics of objects. These attributes help further improve on various downstream tasks such as object detection \cite{obj,ob1,ob3,cobe,crossstitch_cvpr16,neil_iccv13}, action recognition \cite{object_centric_new,perceptual_causality,epic_states1,epic_states,joint_dis, fathi}, image captioning \cite{bt,cap1}, and zero-shot and semi-supervised classification \cite{ob2,causal,grasman,recog_ua,unseen_attrib,shrivastava_eccv12,neil_iccv13}. 
Similar to multi-class classification for objects, initial work for attribute understanding used discriminative models \cite{ob1,ob5}, without understanding attributes. Other works \cite{ob3,joint0,joint,joint1} explored the relation between the same attributes and different objects, to learn visual attributes. Particularly, disentangling object features from attribute features are explored in~\cite{viz_at,viz14}. Although, these works use clustering and probabilistic models to learn the attributes of objects.

\noindent \textbf{Compositional Zero-shot Learning.} Concept of compositional learning was first introduced in Recognition by Parts~\cite{hoff}. Initially,~\cite{redwine} employed this concept for objects and attributes. Unlike zero-shot learning (ZSL), CZSL requires the model to learn to compose unseen concepts from already learned primitive components. \cite{redwine,joint} proposed separate classifiers for primitive components, and merged all into a final classifier. Most prior works use linguistically inspired auxiliary loss terms to regularize training for embedding space, such as:~\cite{ao} models attributes as a linear transformation of objects, ~\cite{symnet} uses rules of symmetry for understanding states, and ~\cite{learn} learns composition and decomposition of attributes hierarchically. Another set of studies uses language priors to learn unseen attribute-object pairs, either in feature space or with multiple networks~\cite{visprod,tmn,adver}. Other recent works use graph structure to leverage information transfer between seen to unseen pairs using Graph Convolutional Networks~\cite{ge,ge2}, and~\cite{rel} uses key-query based attention, along with modular network with message passing for learning relation between primitive concepts.

\vspace{-0.02in}
\section{Object Attribute Disentanglement (OADis)}
\vspace{-0.02in}
Contrary to prior works~\cite{ao,symnet,ge,learn}, we explicitly focus on separating attributes and object features in the visual space. More precisely, TMN~\cite{tmn} uses word embeddings to generate attention layers to probe image features corresponding to a given pair, GraphEmbedding~\cite{ge} exploits the dependency between word embeddings of the labels, and HiDC~\cite{learn} mainly uses word embeddings to compose novel pairs and generate more examples for their triplet loss. To the best of our knowledge, none of the existing works have explored visual feature disentanglement of attributes and objects. We hypothesize that attribute and object visual features can be separated when considering visual feature similarities and differences between image pairs. Composing these disentangled elements help regularize the common embedding space to improve recognition performance. More concretely, we take cues from~\cite{viz_at} and~\cite{redwine,learn}, to learn to compose unseen attribute-object pairs leveraging visual attributes based on auxiliary losses. 

\vspace{-0.02in}
\subsection{Task Formulation}
\vspace{-0.05in}

We follow the conventional Compositional Zero-shot Learning (CZSL) setup, where distinct attribute-object compositions are used at training and testing. Each image $I$ is labeled with $y = y_\text{attr,obj} \in Y$, where $y_\text{attr}$ and $y_\text{obj}$ are respectively the attribute and object label. The dataset is divided into two parts, seen pairs $y^s \in Y^s$ and unseen pairs $y^u \in Y^u$, such that $Y = Y^s \cup Y^u, Y^s \cap Y^u = \emptyset$. Although $y^u = y_\text{attr,obj} \in Y^u$ consists of attribute $y_\text{attr}$ and object $y_\text{obj}$ that are never seen together in training, they are separately seen. We employ the Generalized CZSL setup defined in~\cite{tmn}, which has seen $Y^s$ and unseen pairs $Y^u$ in the validation and test sets as detailed in Table~\ref{tab:data}.
As shown in Figure~\ref{fig:system}, for image $I$, with label \texttt{peeled} \texttt{apple}, we choose two additional images: one with same object and different attribute $\Iob$ (\eg, \texttt{sliced} \texttt{apple}), and another image with same attribute and different object $\Iat$ (\eg, \texttt{peeled} \texttt{orange}). Note that the subscript of image symbol, \eg, $\text{attr}$ in $\Iat$, shows similarity with $I$, whereas superscript denotes seen and unseen sets.

\begin{figure*}
\centering
\includegraphics[width=0.95\linewidth]{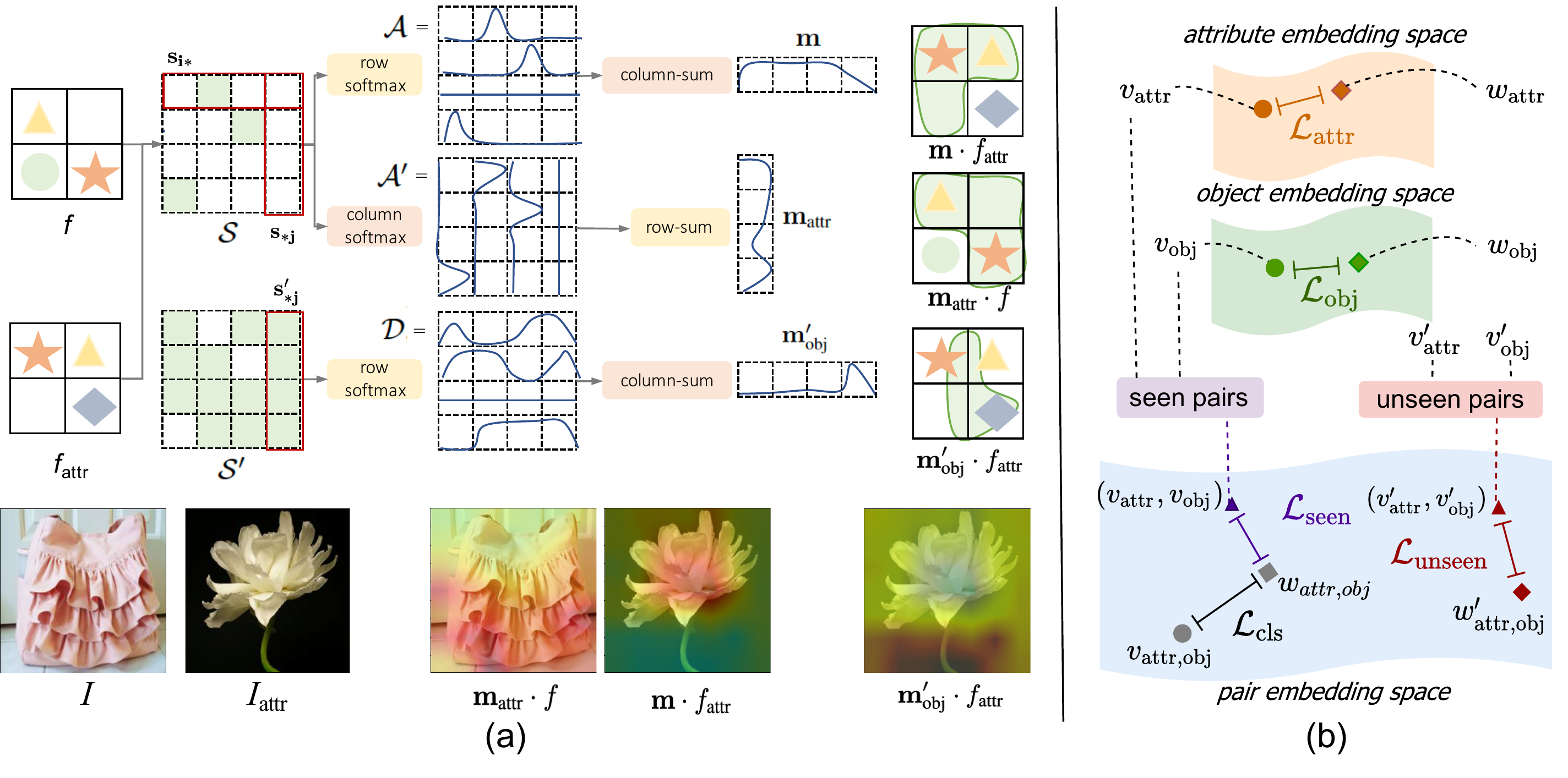}
\vspace{-1.5em}
\caption{(a) Attribute Affinity Module: We compute the cosine similarity between blocks in $f$ and $\fat$ ($S$ in Eq.~\ref{s}), then apply row-wise and column-wise softmax ($\A$ and $\Aprime$), followed by a respective column-sum and row-sum to obtain $\mathbf{m}$ and $\mathbf{\ma}$. $\mathbf{m}$ represents regions where $\fat$ is highly similar to $f$ (hence, we reshape and multiply $\mathbf{m}$ with $\fat$) and $\mathbf{\ma}$ represents regions where $f$ is highly similar to $\fat$ (thus, $\mathbf{\ma}\cdot f$). Similarly, $S^\prime$ represents the regions where feature $\fat$ is not similar to feature $f$ (more details in Section~\ref{sec}). The last row shows real samples and generated attention maps overlayed on images. Give image \texttt{ruffled} \texttt{bag} and \texttt{ruffled} \texttt{flower}, we show that attribute \texttt{ruffle} is highlighted in the center $\mathbf{\ma} \cdot f$ and $\mathbf{m} \cdot \fat$. Whereas, $\mathbf{\mop} \cdot \fat$ shows the dissimilar regions of $\Iat$ w.r.t $I$. (b) Shows the three embedding spaces learnt with different losses. Same notation is used as Figure~\ref{fig:system}.} 
\label{fig:emb_space}
\vspace{-1.3em}
\end{figure*}%

\subsection{Disentangling Visual Features}\label{sec}
\vspace{-0.3em}
We extract image and label embedding features from pre-trained networks (ResNet~\cite{resnet18} and GloVe~\cite{glove}). As seen in Figure~\ref{fig:system}, we use \textit{Image Encoder (IE)} and \textit{Object Conditioned Network (OCN)}, for image and word embedding features respectively. Similar to~\cite{ao}, we use \textit{Label Embedder (LE)} as an additional FC-Layer for the image feature. \textit{LE} and \textit{OCN} learn image and word embeddings and embed those in a common pair embedding space. Next, visual similarity between $I$ and $\Iob$ is computed using \textit{Object Affinity Network}, which extracts visual features for object, $\vo$. Whatever is not similar is considered dissimilar. Hence, visual features of $\Iob$ that are least similar to visual features of $I$ are considered as the attribute feature $\vap$ in $\Iob$, which is \texttt{sliced} in this example. Similarly, \textit{Attribute Affinity Network} takes $I$ and $\Iat$, and extracts visual similarity feature $\va$ for \texttt{peeled}, and dissimilar visual features of $\Iat$, as object feature $\vop$ for \texttt{orange}. The disentangled features are then used to compose seen and unseen pairs. We discuss the details in the following sections:

\smallskip
\noindent \textbf{Image Encoder (IE).} We use the second last layer before AveragePool of an ImageNet-pretrained ResNet-18~\cite{resnet18,imagenet} to extract features for all images. \textit{IE} is a single convolutional layer that is shared across images $I$, $\Iat$ and $\Iob$ to generate their image features, represented as $f$, $\fat$ and $\fob$ respectively, where each $f \in \mathbb{R}^{n \times 49}$ and $n$ is the output dimension of IE.

\smallskip
\noindent \textbf{Label Embedder (LE).}  Inspired by~\cite{ao}, our \textit{LE} inputs spatial feature from ResNet~\cite{resnet18}, AveragePools and passes through a linear layer to extract final feature $\vao$ for pair embedding, which has same dimension as the word embedding final feature $\wao$, extracted from \textit{Object Conditioned Network (OCN)} (Figure~\ref{fig:system}). This is the main branch, and is used for input image $I$ only.

\smallskip
\noindent \textbf{Object Conditioned Network (OCN).}  This takes word embeddings of attribute $emb_\text{attr}$ and object $emb_\text{obj}$, concatenates the features and passes through multiple layers. Object-conditioned is named because a residual connection for the object feature is concatenated with the final attribute feature, and the output feature is $\wao \in  Y$. We discuss the motivation for this in Section~\ref{abl}.

\smallskip
\noindent \textbf{Cosine Classifier (CosCls).} Analogous to compatibility function used in~\cite{ge,ge2}, we use cross-entropy along with cosine similarity to get the final score for each pair. For visual features $\vao$ (from \textit{LE}), and composed word embeddings $\wao$ (from \textit{OCN}), \textit{CosCls} provides logits for an image $I$. For instance, let us assume $v:X \rightarrow Z$ and $w: Y \rightarrow Z$. $Z$ is the common embedding space for word embeddings $w$ and visual embeddings $v$. Then classifier unit \textit{CosCls} gives the score for label $y \in Y^s$ is $C$:
\vspace{-0.75em}
\begin{gather}
h(v,w)=\cos(v,w)=\delta \cdot \frac{v^T w}{\left\|v\right\|\left\|w\right\|}\\
C(v,w)=\frac{e^{ h(v, w)}}{\sum_{y \in Y^s} e^{ h(v, y)}}
\end{gather}
where $\delta$ is the temperature variable. Each loss function uses same CosCls score evaluator, with different inputs. 

\smallskip
\noindent \textbf{Object and Attribute Similarity Modules.} Our main contribution is the proposed affinity modules and compositional losses. Inspired by image captioning~\cite{attn,vis_Sem, imram}, OADis uses image similarities and differences to identify visual features corresponding to attributes and objects. \textit{Object Affinity Network (OAN)} uses $f$ and $\fob$, whereas \textit{Attribute Affinity Network (AAN)} uses $f$ and $\fat$. For brevity, we explain the \textit{AAN}, while the OAN follows the same architecture. Reminded that both $f$ and $\fat \in \mathbb{R}^{n \times 49}$.

Similar to~\cite{attn2}, which computes attention between word concepts with corresponding visual blocks, we compute attention between two images $I$ and $\Iat$. Since both images have the same attribute, \ie, \texttt{peeled}, our affinity network learns visual similarity between the images, which represents the attribute. 
Similarity matrix $\mathcal{S}$ is the cosine similarity between  $f$ and $\fat$, such that $\mathcal{S}\in\mathbb{R}^{49 \times 49}$
as: 
\vspace{-.1in}
\begin{equation} \label{s}
\mathcal{S}=\frac{f^{T} \fat}{\left\|f\right\|_{2}\left\|{\fat}\right\|_{2}}
\vspace{-.1in}
\end{equation}
where element $s_{ij}$ represents the similarity between $i^\text{th}$ element of $f$ with $j^\text{th}$ element of $\fat$. Moreover, let $\mathbf{s_{i*}}$ and $\mathbf{s_{*j}}$ represent the $i^\text{th}$ row and $j^\text{th}$ column of $\mathcal{S}$ respectively. Then, $\mathbf{s_{i*}}$ captures the similarity of all the elements in $\fat$ with respect to $i^\text{th}$ element of $f$. To know the most similar element among $\fat$ with respect to $i^\text{th}$ element of $f$, we can take a row-wise \texttt{softmax} over $\mathcal{S}$. Similarly, for $j^\text{th}$ element of $\fat$, column $\mathbf{s_{*j}}$ represents the similarity with all the elements of $f$. Using a column-wise \texttt{softmax}, we can interpret the most similar and least similar element of $f$ with respect to $j^\text{th}$ element of $\fat$, as shown in Figure~\ref{fig:emb_space}. Therefore, by applying column-wise and row-wise \texttt{softmax}, we get two matrices, $\A$ and $\Aprime$ ($\A, \Aprime \in \mathbb{R}^{d\times d}, d=49$),
\vspace{-.1in}
\begin{gather}
    \A_i =  \frac{e^{\lambda \mathbf{s_{i*}}}}{\sum_{j=1}^d e^{\lambda s_{ij}}} \quad \text{and} \quad
    \Aprime_j =  \frac{e^{\lambda \mathbf{s_{*j}}}}{\sum_{i=1}^d e^{\lambda s_{ij}}},
 \end{gather} where $\lambda$ is the inverse temperature parameter.
We compute row and column sum for $\A$ and $\Aprime$ respectively, to get final similarity maps, $\mathbf{m}$ and $\mathbf{\ma}$,
\vspace{-.1in}
\begin{gather}
    m_j = \sum_{i = 1}^{d}\A_{ij}\quad \text{and} \quad {\ma}_i = \sum_{j = 1}^{d}\Aprime_{ij}.
 \end{gather}

Similarly, the difference between these two images $f$ and $\fat$ is the object label, $\yo$. Hence, we use the negative of $\mathcal{S}$ as the image difference, denoted as $\Sprime$. Then, difference of $\fat$ with respect to $f$ would be row-wise \texttt{softmax} of difference matrix, denoted by $\D$. Hence, by performing column-sum over $\D$, we get difference map, $\mathbf{\mop}$,
\vspace{-.1in}
\begin{gather}
     \mathcal{D}_j =  \frac{e^{\gamma \mathbf{s^\prime_{*j}}}}{\sum_{i=1}^d e^{\gamma s^\prime_{ij}}} \quad \text{and} \quad
     {\mop}_i = \sum_{j = 1}^{d}\D_{ij}.
 \end{gather} The final disentangled features for attribute $\va$ and object $\vop$, for both AAN and OAN, can be computed as:
 \vspace{-.5em}
 \begin{equation}
\begin{split}
      \va = \mathbf{m} \cdot \fat + \mathbf{\ma} \cdot f \quad &\text{and} \quad
    \vop = \mathbf{\mop} \cdot \fat\\
      \vo = \mathbf{m} \cdot \fob + \mathbf{\mo} \cdot f \quad &\text{and} \quad
     \vap = \mathbf{\map} \cdot \fob.
\end{split}
\end{equation}
More details using a toy example can be seen in Figure~\ref{fig:emb_space}.
 Using concatenation of $\va$ and $\vo$ along with a single Linear layer, composes the pair \texttt{peeled} \texttt{apple}, represented by $(\va,\vo)$. Similarly, the disentangled visual features $\vap$ and $\vop$, are used to compose unseen pair \texttt{sliced} \texttt{orange}, and is represented as $(\vap,\vop)$.

\vspace{-0.02in}
\subsection{Embedding Space Learning objectives}
\vspace{-0.03in}
As shown in Figure~\ref{fig:emb_space}b, we learn three embedding spaces: (1) attributes space, (2) object space, and (3) attribute-object pair space. The attribute and object spaces are used for disentangling the two, whereas pair embedding is used for final pair composition and inference. OADis has separate loss functions for disentangling and composing. All loss functions are expressed in terms of \textit{CosCls} defined previously.

The loss function for main branch, $\lcls$ uses combined visual feature $\vao$ from \textit{LE} and word embedding feature $\wao$ from \textit{OCN}. $\lcls$ is used for the pair embedding space. Similarly, $\lat$ and $\lob$ are used to learn the visual attribute and object feature, in their respective embedding spaces. $\lat$ pushes the visual feature of attribute, closer to the word embedding. $\lob$ does the same for objects in object embedding space Figure~\ref{fig:emb_space}b. These losses cover the concept of disentanglement, and can be represented as:
 \vspace{-0.5em}
 \begin{equation} 
\begin{split}
\lcls &= C(\vao,\wao)\\
\lat &= C(\va,\wa);\,\,
\lob = C(\vo,\wo)\\
\end{split}
\end{equation}

For composition, we use $\lse$ and $\lun$. Among the images seen ($I$, $\Iat$, and $\Iob$), disentangled features $\vo$ and $\va$, composes the same pair as $(\va,\vo)$, which we refer to as the seen composition. Note that $(\va,\vo)$ is different from $\vao$, as the former is hallucinated feature with combination of disentangled attribute and object visual features, and latter is the combined visual feature extracted with \textit{LE}. Here, we use $\lse$ loss which takes the composition of disentangled features and learns to put the composition closer to $\wao$. Moreover, the dissimilarity aspect from \textit{OAN} and \textit{AAN} extracts $\vap$ and $\vop$, which composes an unseen pair $(\vap,\vop)$. We use $\lun$ as unseen loss since the hallucinated composition is never seen among $I$ , $\Iat$, and $\Iob$. 

\vspace{-0.5em}
\begin{equation} 
\begin{split}
\lse &= C((\va,\vo),\wao)\\
\lun &= C((\vap,\vop),\waop)\\
 \end{split}
 \end{equation}
The combined loss function $\mathcal{L}$ is minimized over all the training images, to train OADis end-to-end. The weights for each loss ($\mathcal{\alpha}$) are empirically computed:

\vspace{-0.5em}
\begin{equation*}
\begin{split}
 \mathcal{L} &= \lcls + \alpha_1 \lat + \alpha_2 \lob + \alpha_3 \lse  + \alpha_4 \lun.
 \end{split}
 \end{equation*}
 \vspace{-1.3em}

\begin{table}[t]
\caption{This table shows dataset splits. $Y^s$ and $Y^u$ are seen and unseen compositions respectively. We propose a new benchmarck, VAW-CZSL~\cite{khoi}, which has more than 10$\times$ compositions in each split compared to other datasets.}
 \label{tab:data}
  \centering
  \footnotesize
  \vspace{-0.1in}
   \renewcommand{\tabcolsep}{3pt}
 \begin{tabular}{@{}lccccc@{}}
 \toprule
 &\multicolumn{3}{c}{Train set} & Val set & Test set \\
\cmidrule{2-4}
\cmidrule(lr){5-5}
\cmidrule{6-6}
  Datasets & attr. & obj. & $Y^s$ & $Y^s / Y^u$ & $Y^s / Y^u$ \\
 \midrule
 MIT-states~\cite{mitstates} & 115 & 245 & 1262  & 300 / 300 &400 / 400\\
 UT-Zappos~\cite{utzappos} &16 & 12 & 83 & 15 / 15 & 18 / 18 \\
VAW-CZSL~\cite{khoi} & 440 & 541 & 11175 & 2121 / 2322 &	2449 / 2470	\\
 \bottomrule
 \end{tabular}
 \vspace{-0.1in}
 \end{table}

\begin{table*}[t]
\caption{We show results on MIT-states~\cite{mitstates} and UT-Zappos~\cite{utzappos}. Following~\cite{ge,tmn}, we use AUC in \% between seen and unseen compositions with different bias terms, along with Val, Test, attribute and object accuracy. HM is Harmonic Mean. OADis consistently outperforms on most categories with significant increment.} \label{tab:res1}
  \centering
  \small
  \vspace{-0.1in}
  \renewcommand{\tabcolsep}{5pt}
 \resizebox{0.95\textwidth}{!}{
 \begin{tabular}{@{}lccccccccccccccc@{}}
  \toprule
 & \multicolumn{7}{c}{MIT-States}
   & \multicolumn{7}{c}{UT-Zappos} \\
\cmidrule{2-8}
\cmidrule(l){9-16}
Model & Val@1 & Test@1 & HM & Seen & Unseen & Attribute & Object && Val@1 & Test@1 & HM & Seen & Unseen & Attribute & Object\\
\midrule
AttrOpr~\cite{ao} & 2.5 & 2.0 & 10.7 &16.6 & 18.4 & 22.9	&24.7 &&	29.9 &	22.8 &  38.1 & 55.5 & 54.4	& 38.6 & 70.0\\	
LabelEmbed+~\cite{ao}	&3.5	&	2.3&	11.5	&16.2	&21.2	&25.6	&27.5 && 	35.5 &	22.6&	37.7 &	53.3 &	58.6 &	40.9 & 69.1\\
TMN~\cite{tmn}	&3.3	&	2.6	&11.8	&22.7	&17.1	&21.3	&24.2 &&35.9 & 28.4 & 44.0 & 58.2& 58.0 &40.8& 68.4\\
Symnet~\cite{symnet}	&4.5	&	3.4		&13.8	&24.8&	20.0	&26.1	&25.7 &&	27.4 &	27.7 &	42.5 &	56.7	& 61.6 & 44.0 & 70.6\\
CompCos~\cite{ge2}	&6.9	&4.8	&16.9	&26.9	&24.5	&28.3	&31.9 & & \textbf{40.8} & 26.9 &  41.1 & 57.7 & 62.8	& 43.3 & 73.0\\
GraphEmb~\cite{ge}&	7.2	&	5.3& 18.1&	28.9&	25.0	&27.2	&32.5 & & 33.9 &  24.7& 38.9	& 58.8	& 61.0	& 44.0 &	72.6\\
\midrule
\textbf{OADis}&	\textbf{7.6} & \textbf{5.9} & \textbf{18.9} &\textbf{31.1} &\textbf{25.6} & \textbf{28.4} & \textbf{33.2} & & \textbf{40.8} & \textbf{30.0} & \textbf{44.4} & \textbf{59.5} & \textbf{65.5} & \textbf{46.5} & \textbf{75.5}\\
\bottomrule
 \end{tabular}}
 \vspace{-0.1in}
 \end{table*}
\vspace{-1em}
\section{Experiment}
\noindent\begin{table}[t]
\caption{We show results on VAW-CZSL. Since it is a much more challenging dataset, with significantly large number of compositions, to discriminate performance among different baseline, we show top-3 and top-5 AUC (in \%) for Val and Test sets.}
\vspace{-0.1in}
 \label{tab:res2}
  \centering
\footnotesize
\renewcommand{\tabcolsep}{2.5pt}
\renewcommand{\arraystretch}{1.05}
\resizebox{\linewidth}{!}{%
\begin{tabular}{@{}lcccccccccccc@{}}
  \toprule
 & \mc{2}{c}{Val.\ Set} & \mc{2}{c}{Test Set} \\ 
\cmidrule(lr){2-3}
\cmidrule(lr){4-5}
Model  & V@3 & V@5 & V@3 & V@5 & HM & Seen & Unseen & Attr.\ & Obj.\ \\
\midrule
AttrOpr~\cite{ao} & 1.4 & 2.5 &	1.4 & 2.6 &	9.1 & 16.4 &	11.7 & 13.7 & 34.9 \\	
LabelEmbed+~\cite{ao} &	1.5 & 2.8 &	1.6 & 2.8 &	9.8 &	16.2 & 13.2 & 13.4 & 35.1 \\
Symnet~\cite{symnet} & 2.3 & 3.9 & 2.3 & 3.9 & 12.2 &	19.1 & 15.8 & \textbf{18.6} & 40.9 \\
TMN~\cite{tmn} & 2.2 & 3.9 & 2.3 & 4.0 & 11.9 & 19.9 & 15.4 & 15.9 & 38.3 \\
CompCos~\cite{ge2} & 3.1 & 5.6 & 3.2 & 5.6 & 14.2 & 23.9 & 18.0 & 16.9 & 41.9 \\
GraphEmb~\cite{ge} & 2.7 & 5.3 & 2.9 & 5.1 & 13.0 &	23.4 & 16.8 & 16.9 & 40.8 \\
\midrule
\textbf{OADis}&	\textbf{3.5} & \textbf{6.0}	& \textbf{3.6} & \textbf{6.1} & \textbf{15.2} & \textbf{24.9} & \textbf{18.7} &	17.5 & \textbf{43.3} \\
\bottomrule
\end{tabular}
}
\vspace{-0.15in}
\end{table}
 
\vspace{-2em}
\subsection{Datasets and Metrics}
\vspace{-0.4em}
We show results on three datasets: MIT-states~\cite{mitstates}, UT-Zappos~\cite{utzappos}, and a new benchmark for evaluating CZSL on images of objects in-the-wild, referred as VAW-CZSL. VAW-CZSL is created based on images with object and attribute labels from the VAW dataset~\cite{khoi}. Both MIT-states~\cite{mitstates} and UT-Zappos~\cite{utzappos} are common datasets used for this task in previous studies. MIT-states covers wide range of objects (\ie, laptop, fruits, fish, room, \etc) and attributes (\ie, mossy, dirty, raw, \etc), whereas UT-zappos has fewer objects (\ie, shoes type: boots, slippers, sandals) and fine-grained attributes (\ie, leather, fur, \etc). 

\smallskip
\noindent \textbf{Proposed New Benchmark.} While experimenting with MIT-states~\cite{mitstates} and UT-Zappos~\cite{utzappos}, we found several shortcomings with these datasets and discovered issues across all baselines using these datasets:
\begin{itemize}[noitemsep, nolistsep, labelindent=0pt, itemindent=0em, leftmargin=*]
\setlength{\itemsep}{1pt}
  \setlength{\parskip}{0pt}
  \setlength{\parsep}{0pt}
    \item Both datasets are small, with a maximum of 2000 attribute-object pairs and 30k images, leading to overfitting fairly quickly.
    \item Random seed initialization makes performance fluctuate significantly (0.2-0.4\% AUC). Moreover,~\cite{causal} found 70\% noise in human-annotated labels on MIT-States~\cite{mitstates}.
    \item A new dataset C-GQA was introduced in~\cite{ge}, but the dataset is still small and we found a lot of discrepancies (kindly refer to the suppl.).
\end{itemize}

To address these limitations, we propose a new benchmark \textbf{VAW-CZSL}, a subset of VAW~\cite{khoi}, which is a multi-label attribute-object dataset. We sample one attribute per image, leading to much larger dataset in comparison to previous datasets as shown in Table~\ref{tab:data} (details in the suppl.). 

\smallskip
\noindent \textbf{Evaluation.} We use Generalized CZSL setup, defined in~\cite{tmn}, with dataset statistics presented in Table~\ref{tab:data}. As observed in prior works~\cite{ge,tmn}, a model trained on a set of labels $Y^s$, does not generalize well on unseen pairs $Y^u$. Therefore,~\cite{ge,tmn} use a scalar term for overcoming the negative bias for unseen pairs. We use the same evaluation protocol, which computes Area Under the Curve (AUC) (in \%) between the accuracy on seen and unseen compositions with different bias terms~\cite{tmn}. Larger bias term leads to better results for unseen pairs whereas smaller bias leads to better results for seen pairs. Harmonic mean is reported, to balance the bias. We also report the attribute and object accuracy for unseen pairs, to show improvement due to visual disentanglement of features. Our new benchmark subset for VAW~\cite{khoi}, follows the similar split as other datasets. In addition, we conduct all experiments with image augmentation for all methods (discussed in Section~\ref{abl}).

\vspace{-0.2em}
\subsection{Results and Discussion}
\label{ab}
\vspace{-0.2em}

\noindent \textbf{Baselines.} We compare with related recent and prominent prior works: AttrOp~\cite{ao}, LabelEmbed+~\cite{ao}, TMN~\cite{tmn}, Symnet~\cite{symnet}, CompCos~\cite{ge2} and GraphEmb~\cite{ge}. We do not compare with BMP~\cite{rel}, since it uses the
concatenation of features from all four ResNet blocks (960-d features), resulting in higher input features and the number of network parameters than all other setups. Moreover, GraphEmb~\cite{ge} is state-of-the-art; hence, comparing with that makes our work comparable to other baselines that~\cite{ge} already outperforms. To be consistent, we state the performance of all models (including GraphEmb~\cite{ge}) using frozen backbone ResNet without fine-tuning the image features, and using GloVe~\cite{glove} for the object and attribute word embeddings. Before passing through backbone, training images are augmented with horizontal flip and random crop. Compared to other baselines, OADis uses convolutional features rather than AvgPooled, since it is easier to segregate visual features in the spatial domain for attributes and objects. Moreover, other studies~\cite{ge, ge2} have also used additional FC layers on top of \textit{IE}, which we argue makes it fair for us to use pre-pooled features for OADis.

\begin{figure*}[t]
\centering
\includegraphics[width=0.9\linewidth]{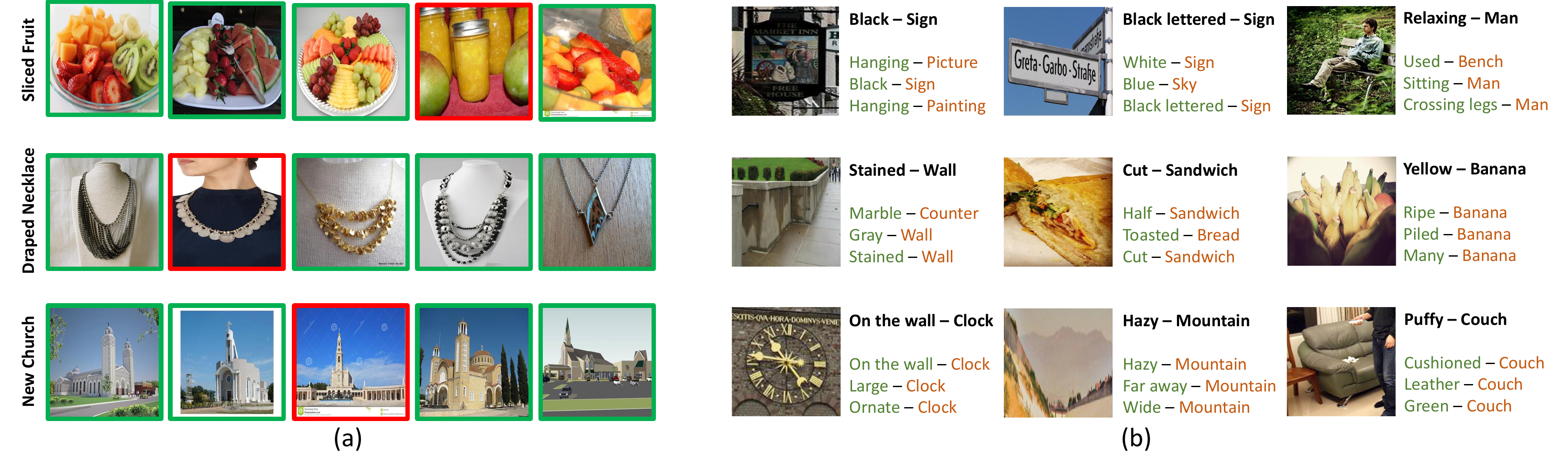}
\vspace{-0.1in}
\caption{\textbf{Qualitative Results}: We show the nearest neighbors using the hallucinated unseen composition features for MIT-states and UT-Zappos. Although, all the neighbors are not correct (represented with \red{red} outline), they look very similar to true class labels: (a) First row: \red{pureed fruit}, Second row: \red{engraved coin}, Third row: \red{huge tower}. (b) We show top-3 predictions for images in VAW-CZSL. } 
\label{fig:mit}
\vspace{-0.05in}
\end{figure*}

\begin{table*}
\caption{We quantitatively show that the proposed architecture and different losses help in disentanglement and composition of unseen pairs. The experiments are conducted on MIT-States~\cite{mitstates}, where change in accuracy is shown with \green{green} and \red{red} based on increment or decrement respectively from the previous row. A dash (-) represents no change more than ($\pm$ 0.1). Refer to Section~\ref{ab} for details.}
 \label{tab:losses}
  \centering
  \small
  \vspace{-0.1in}
  \renewcommand{\tabcolsep}{5pt}
  \resizebox{0.8\textwidth}{!}{
 \begin{tabular}{@{}lllllll@{}}
  \toprule
Losses & Val AUC@1 & Test AUC@1  & Seen & Unseen & Attribute & Object \\
\midrule
$\lcls$ & 7.24 & 5.43 & 29.92   & 25.33 & 28.03 & 33.10 \\
$\lcls$ + $\lat$ &  - &	- &	31.09 \scriptsize\green{(+2.0)} &	- &	28.30 \scriptsize\green{(+0.3)} &	-\\
$\lcls$ + $\lob$ & - & - &	- &	25.50 \scriptsize\green{(+0.2)} &	- &	33.38 \scriptsize\green{(+0.2)}\\
$\lcls$ + $\lat$ + $\lob$ & 7.49 \scriptsize\green{(+0.2)} &	5.73 \scriptsize\green{(+0.2)} &	- &	- &	28.50 \scriptsize\green{(+0.2)} &	- \\
$\lcls$ + $\lat$ + $\lob$ + $\lse$ & -	& 5.44 \scriptsize\red{(-0.5)}	& 31.21	\scriptsize\green{(+0.2)} & - &	28.18 \scriptsize\red{(-0.4)} &	-\\
$\lcls$ + $\lat$ + $\lob$ + $\lun$ & - &	5.73 \scriptsize\green{(+0.3)} & - &	25.80 \scriptsize\green{(+0.4)} &	28.51 \scriptsize\green{(+0.4)} &	-\\
$\lcls$ + $\lat$ + $\lob$ + $\lse$ + $\lun$ & 7.62 \scriptsize\green{(+0.2)}	& 5.94 \scriptsize\green{(+0.2)} &	31.64 \scriptsize\green{(+0.4)} &	25.60 \scriptsize\red{(-0.2)} &	28.51 &	33.20 \\
\bottomrule
 \end{tabular}}
 \vspace{-0.15in}
 \end{table*}

\begin{table}
\caption{Results with different networks for word-embeddings. Object-conditioning with attribute performs the best, and is therefore used for OADis (Section~\ref{abl}).}
 \label{tab:word}
  \centering
  \small
  \vspace{-0.1in}
\resizebox{0.7\columnwidth}{!}{
 \begin{tabular}{@{}lccc@{}}
\toprule
 & Linear & MLP & Obj-cond.\ Network \\
\midrule
Val@1 &6.6& 7.0 & 	\textbf{7.6}\\
Test@1 & 5.0	& 5.2 & \textbf{5.9} \\
\bottomrule
 \end{tabular}}
 \vspace{-0.13in}
 \end{table}
 
\noindent \textbf{Results on MIT-States.}  MIT-states has considerable label noise~\cite{causal}, but still is a standard dataset for this task. We show significant improvement on this dataset (reported in Table~\ref{tab:res1}), from previous state-of-the-art GraphEmb, which has 7.2 Val AUC and 5.3 Test AUC.
Note that we do not report GraphEmb results with fine-tuning backbone, as we find it incomparable with other baselines that did not incorporate fine-tuning as part of their proposed methods. Overall, our model performs significantly better than GraphEmb on all metrics.

\noindent \textbf{Results on UT-Zappos.} Similar improvement trends hold for UT-Zapopos as well (see Table~\ref{tab:res1}). Although, as explained for GraphEmb, it is difficult to balance the best performance for Val and Test set in this dataset. The problem is that 7/36 ($\sim$20\%) attributes in Test set do not appear in Val set. Hence, improving Val set AUC, does not necessarily improve Test AUC for UT-Zappos. Similar trend can be seen for other baselines: CompCos has best Val AUC, but does not perform well on Test set, compared to TMN and Symnet. Even GraphEmb in their final table show the frozen backbone network has much lower performance than TMN. However, OADis performs well on UT-Zappos overall, with $\sim$4.0 improvement for Val and Test AUC, HM, unseen and object accuracy.

\noindent \textbf{Results on VAW-CZSL.}
Our model performs well on VAW-CZSL, and is consistently better than other methods across almost all metrics. As shown in Table~\ref{tab:data}, VAW-CZSL has $\sim$6-8 times more pairs in each split than MIT-States, which shows how challenging the benchmark is. Due to top-1 AUC being too small to quantify any learning and comparing between methods, we report top-3 and top-5 AUC instead. This is also because objects in-the-wild tend to depict multiple possible attributes; hence, evaluating only the top-1 prediction is insufficient. We provide qualitative results of how our model makes object-attribute composition prediction on VAW-CZSL in the suppl.

\begin{figure}
\centering
\includegraphics[width=0.95\linewidth]{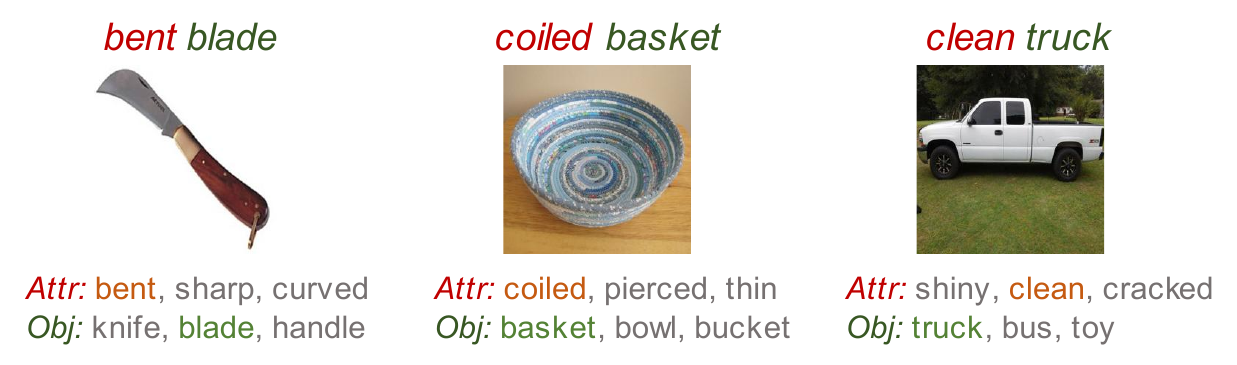}
\vspace{-0.1in}
\caption{Qualitative results showing top 3 attributes and objects from test images, using prototype disentangled features computed on training data.}
\label{fig:dis}
\vspace{-0.2in}
\end{figure}

\smallskip
\noindent \textbf{Is disentangling and hallucinating pairs helpful?}
Prior works rely heavily on word embeddings for this task, but to improve the capabilities of visual systems, it is imperative to explore what is possible in the visual domain. We do an extensive study to understand if our intuition aligns with OADis (Table~\ref{tab:losses}). Here are some takeaways:
\begin{itemize}[noitemsep,left=0pt]
  \setlength{\itemsep}{1pt}
  \setlength{\parskip}{0pt}
  \setlength{\parsep}{0pt}
  \vspace{-0.5em}
    \item Using only $\lcls$, we get a benchmark performance based on the architectural contributions, such as \textit{LE} and \textit{ONC}. When $\lat$ is added, significant performance boost for attribute accuracy can be seen in Table~\ref{tab:losses}. 
    \item Adding object loss $\lob$ with $\lcls$, makes object accuracy better but no change in Val and Test AUC. This indicates the need of both losses to balance the effects. Using both $\lat$ and $\lob$ gives improvement in all measures.
    \item Adding $\lse$ results in boost for seen AUC, but drop in Test AUC, which has unseen pairs along with seen pairs. Using unseen loss $\lun$ leads to increase in both Test and attribute accuracy.
    \item Finally adding unseen composition loss $\lun$ along with seen loss  $\lse$, the model improves on most metrics. Each loss plays a role and regularizes effects from other losses. 
\end{itemize}
\vspace{-0.5em}
\noindent \textbf{Is visual disentangling actually happening?}
Visual disentanglement in feature space is challenging to visualize since: (a) parts of an image for attributes and objects are hard to distinguish, as \textit{attributes are aspects of an object}; (b) OADis is end-to-end trained with losses to disentangle features for attribute and object embeddings, which is separate from pair embedding space. Inspired by \cite{ao,symnet}, we show a few qualitative results in Figure~\ref{fig:dis}. Using all training images, prototype features $\mathcal{V_\text{attr}}$ for each attribute can be computed by averaging features for all images containing that attributes $\va$ using \textit{AAN}. Similarly, with \textit{OAN}, prototype object features are also computed. For each test image, we find top-3 nearest neighbors from these prototype features (Figure~\ref{fig:dis}). Hence, the disentangled prototype features of attributes and objects are used for classifying unseen images. Note that results reported in Table 1 use pair embedding space for attribute and object classification, whereas here we use auxiliary attribute and object embedding spaces (in Figure~\ref{fig:emb_space}b) for the same task. If disentanglement features are not robust, then composition features will also not be efficient. We also show that using the composition of disentangled features for unseen pairs, relevant images from the test set can be found in suppl. 

\smallskip
\noindent \textbf{Limitations.} Despite OADis outperforming prior works on all benchmarks, we still notice some outstanding deficiencies in this problem domain. First, similar to~\cite{ge}, OADis often struggles on images containing multiple objects, where it does not know which object to make prediction on. One possible solution is to utilize an object-conditioned attention that allows the model to focus and possibly output attribute for multiple objects. Second, from qualitative studies on VAW-CZSL, we notice there are multiple cases where OADis makes the correct prediction but is considered incorrect by the image label. This is due to the fact that objects in-the-wild are mostly multi-label (containing multiple attributes), which none of the current single-label benchmarks have attempted to address.

\vspace{-0.03in}
\subsection{Ablation Studies} \label{abl}
\vspace{-0.02in}

In this section, we show experiments to support our design choices for OADis. All the ablations are done for MIT-states~\cite{mitstates}, for one random seed initialization, and are consistent for other datasets as well. Empirical results for $\lambda$, $\delta$ and different word embeddings can be found in suppl.

\noindent \textbf{Why Object-Conditioned Network?}
Label Embedder~\cite{ao} uses a linear layer and concatenates word embeddings for attributes and objects. We experiment with other networks: MLP with more parameters with two layers and ReLU and Object-conditioned network that uses a residual connection for object embedding. Our intuition is that same attribute contributes differently to each object, \ie, \texttt{ruffled} \texttt{bag} is very different from \texttt{ruffled} \texttt{flower}. Hence, attributes are conditioned on object. Adding a residual connection for object embeddings to the final attribute embedding helps condition the attribute. We empirically demonstrate that object-conditioning helps in Table~\ref{tab:word} (refer to the suppl.).

\noindent \textbf{To augment or not to augment?}
 Augmentation is a common technique to reduce over-fitting and improve generalization. Surprisingly, prior works do not use any image augmentation. OADis without augmentation gives 6.7\% AUC on Val and 5.1\% AUC on Test set for MIT-states. Hence, we use augmentation for OADis and re-implemented rest of the baselines in Table~\ref{tab:res1}, showing that augmentation helps improving all methods $\sim$1.0-1.5\% AUC. We use horizontal flip and random crop as augmentation.  

\subsection{Qualitative results}
\vspace{-0.02in}
To qualitatively analyze our hallucinated compositions, we perform a nearest neighbor search on all three datasets. We pick the unseen compositions composed using the disentangled features, and find their top-5 nearest neighbors from the validation and test set. Figure~\ref{fig:mit}(a) illustrates a few of our results. Note that these pairs are never seen in training. Based on the hallucinated compositions of disentangled attributes and objects, we are able to retrieve samples from these unseen compositions.

In Figure~\ref{fig:mit}(b), we show the top-3 predictions of OADis on VAW-CSZL. Column 1 shows results for seen, and columns 2 and 3 show unseen compositions, with the ground-truth label on top (bold black). In all examples, our top-3 predictions describe the visual content of the images accurately, even though in many cases the ground-truth label is not predicted in top-1. For column 3, we purposely show examples where our model predictions totally differ from the ground-truth label, but still correctly describe the visual information in each image. Similar to~\cite{ge}, this explains the multi-label nature of object-attribute recognition, and why we report top-3 and top-5 metrics for the VAW-CZSL benchmark.

\vspace{-0.5em}
\section{Conclusion}
\vspace{-0.4em}
In this work, we demonstrated the ability to disentangle object and attribute in the visual feature space, that are used for hallucinating novel complex concepts, as well as regularizing and obtaining a better object-attribute recognition model. Through extensive experiments, we show the efficacy of our method, and surpass previous methods across three different benchmarks. In addition, we also propose a new benchmark for the compositional zero-shot learning task with images of objects in-the-wild, which we believe can help shift the focus of the community towards images in more complex scenes. Finally, we also highlight limitations of our work, including the notable problem of multi-label in object attributes, which we hope would encourage future works to start tackling CSZL for more realistic scenarios.

\smallskip
\noindent\textbf{Acknowledgements.} This work was supported by the Air Force (STTR awards FA865019P6014, FA864920C0010), DARPA SAILON program (W911NF2020009) and gifts from Adobe collaboration support fund.

{\small
\bibliographystyle{ieee_fullname}
\bibliography{egbib}
}

\clearpage
\appendix
\section*{Appendix}
\label{sec:appendix}
\section{Dataset issues of C-GQA \cite{ge}}
\noindent GraphEmb~\cite{ge} proposes a new benchmark for compositional zero-shot learning. However, there are some issues with the dataset have been raised on their official github page~\cite{git1,git2}. These issues are related to (1) the attribute-object pairs being placed into the incorrect train, validation, and test subset, and (2) there are missing images for a decent amount of pairs (20\%), which could potentially affect the final experiment results. Due to~\cite{ge} being unable to provide a corrected version of the dataset in time before the CVPR 2022 deadline, we were unable to run any experiments for C-GQA. Post the deadline, we did run some preliminary results where our method outperformed GraphEmb~\cite{ge}. Although, a major issue we observed was for OADis, C-GQA~\cite{ge} training set did not have  similar attributes and objects samples for constructing $\Iat$ and $\Iob$. However, we propose for learning compositional concepts, firstly disentangled concepts must be learnt, and for that, we require $\Iat$ and $\Iob$. Hence, we do not report results on C-GQA for OADis.

\section{Dataset Creation: VAW-CZSL}
\noindent We propose a new benchmark for the compositional zero-shot learning task (CZSL), focusing on images of objects and attributes in the wild that span across a much larger number of categories. We select the VAW dataset~\cite{khoi} to create our benchmark. VAW contains images originally from Visual Genome (thus objects and attributes in the wild). Every image of an object instance contains an object label and one (or possibly multiple) attribute labels. In the followings, we describe our steps in creating the VAW-CZSL benchmark, which shares some similarities with the C-GQA dataset. 

Different from C-GQA, we consider object instances whose bounding boxes are larger than 50 x 50. C-GQA selected instances whose boxes are larger than 112 x 112, which could possibly leave out small, narrow objects that are still recognizable from images. For every object instance, among its possibly multiple attributes label, we keep only one attribute that has the lowest frequency in the dataset (\ie, the uncommon attribute) to be consistent with the standard CZSL benchmark. By keeping the most uncommon attribute and using the top-3 \& 5 evaluation metrics, all methods will be evaluated based on whether they are able to rank this uncommon (but still representative) attribute in its top-3 \& 5 predictions rather than always predicting the most frequent attributes. From this, we follow the similar steps from \cite{ge} to merge plurals and synonyms (e.g., \{\textit{airplane, plane, aeroplane, airplanes...}\}, \{\textit{rock, stone, rocks...}\}). We then keep only those attribute and object categories with frequency greater than 30 to make sure all primitive concepts have a decent amount of data for training and evaluating.

We use images in VAW-training as our training set, and use images in VAW-val and VAW-test for creating the validation and testing splits following the standard generalized benchmark in CZSL. We first merge VAW-val and VAW-test in one set, and follow similar steps mentioned in~\cite{ge} to create a validation and test set of seen and unseen attribute-object pairs. At the end, we remove objects and attributes that no longer appear in the training set. This is because a model that has never seen an attribute (or object) will find it impossible to generalize to unseen pairs containing this attribute (or object). This problem happens with the C-GQA dataset where ~8\% of attribute and ~22\% of object categories do not exist in their training set. More details about dataset can be found in Table~\ref{tab:data_appendix}. The dataset splits are made publicly available at \url{https://github.com/nirat1606/OADis}.
\section{Implementation Details} 
\noindent Following baselines, we use ResNet18~\cite{resnet18} pre-trained on Imagenet~\cite{imagenet} as backbone feature extractor. Since, proposed auxiliary losses leverage image features, we use a single convolutional layer with Batch Normalization, ReLU and dropout for Image embedder with output dimension 1024 and dropout as 0.3. Note that we extract ResNet features before average pool. For word embeddings, we initialize with GLoVe~\cite{glove}. Object  Conditioned network, uses multiple linear layers, first for objects and attributes separately, then for concatenated features. Label embedder takes 1024-$d$ feature, performs AveragePool and finally embeds in a 300-$d$ space. Each loss uses compatibility function, i.e. cosine similarity, followed by cross-entropy loss over the compatibility function. Object similarity and attribute similarity modules also use two linear layers with dropout 0.05. On UT-Zappos, because the dataset is very small, we find using a linear layer (a smaller and simpler module than OCN) with dropout 0.1 results in better performance. We use Adam optimizer with weight decay $5e^{-5}$, and learning rate $2.5e^{-6}$ for the GLoVe embedding. The learning rate for the rest of the model is $3e^{-4}$ on MIT-States, and $1e^{-4}$ on UT-Zappos and VAW-CZSL. We decay the learning rate by 10 at epoch 30 and 40 on MIT-States, at epoch 50 on UT-Zappos, and at epoch 70 on VAW-CZSL. OADis
needs to be trained for 70-150 epochs depending on the dataset, and training time is comparable with other methods (5-7 hours). These implementation details are also provided in our released source code.

\begin{table*}
\caption{Dataset Details: This table shows the statistics for different datasets and their splits. The proposed VAW-CZSL benchmark significantly increases the number of attributes and objects.}
 \label{tab:data_appendix}
  \centering
  \small
  \vspace{-0.1in}
   \renewcommand{\tabcolsep}{3pt}
 \begin{tabular}{@{}lcccc|ccc|cccc@{}}
 \toprule
 & &&Train set &&& Val set &&&Test set \\
\cmidrule(l{5pt}r{4pt}){2-5}
\cmidrule(l{5pt}r{4pt}){6-8}
\cmidrule(l{5pt}r{4pt}){9-11}
  Datasets:& Attr. & Obj. & Seen Pairs. & \# Images &  Seen Pairs & Unseen Pairs  &\# Images  & Seen Pairs & Unseen Pairs & \# Images\\
 \midrule
 MIT-States~\cite{mitstates} & 115 & 245 & 1262 & 30338 & 300 &300 & 10420 &400& 400& 12995\\
 UT-Zappos~\cite{utzappos} &16 & 12 & 83 & 22998 & 15 & 15 & 3214 & 18 & 18 & 2914 \\
 VAW-CZSL~\cite{khoi} & 440 & 541 & 11175 & 72203 & 2121 & 2322 & 9524 &	2449 & 2470	& 10856\\
 \bottomrule
 \end{tabular}
 \vspace{-0.1in}
 \end{table*}

\section{Ablation studies (extension)}
\label{abl_appendix}
\noindent As mentioned in the paper, we show ablation for various other parameters. 
All  the  ablations  are  done  for  MIT-states~\cite{mitstates}, for one random seed initialization, and are consistent for other datasets as well. 
\subsection{Choice of word embeddings}
\label{abl0}
\noindent Prior works~\cite{symnet,ge,ge2} experiment with various kinds of word embeddings. In fact, GraphEmb~\cite{ge} has more advantages over all other baselines, since they use a combination of word embeddings word2vec~\cite{word2vec} and fasttext~\cite{fasttext}, whereas rest of the works use GloVe~\cite{glove} only. To keep the results fair between all methods, we run all the baselines, even GraphEmb~\cite{ge} with only GloVe~\cite{glove}, and report the accuracy in Table 1, in the main paper. Results for using different embedding combinations is shown in Table~\ref{tab:word1}. Overall, since our method uses word embeddings for visual disentanglement, the choice of word embeddings does not impact the performance much. Although, empirically, we found our model performs best when GloVe embeddings are used.
 \begin{table}
\caption{Results with pre-trained word-embeddings. GloVe~\cite{glove} performs the best, and is therefore used for OADis. (Sec~\ref{abl0})}
 \label{tab:word1}
  \centering
  \small
  \vspace{-0.1in}
\resizebox{0.85\columnwidth}{!}{
 \begin{tabular}{@{}lcc@{}}
 \toprule
 Word Embs & Val AUC@1 & Test AUC@1 \\
\midrule
\textbf{Glove}	&\textbf{7.6}&	\textbf{5.9} \\
Fasttext	&7.4&	5.3 \\
Word2vec	&7.5	&5.4 \\
Glove+fasttext	&7.4	&5.5 \\
Glove+word2vec	&7.5&	5.6 \\
Fasttext+word2vec	& 7.4 & 5.6 \\
\bottomrule
\end{tabular}}
\end{table}
\vspace{-1.2em}
\subsection{Object-conditioned network}
\label{abl1}
\begin{figure}
\centering
\includegraphics[width=0.85\linewidth]{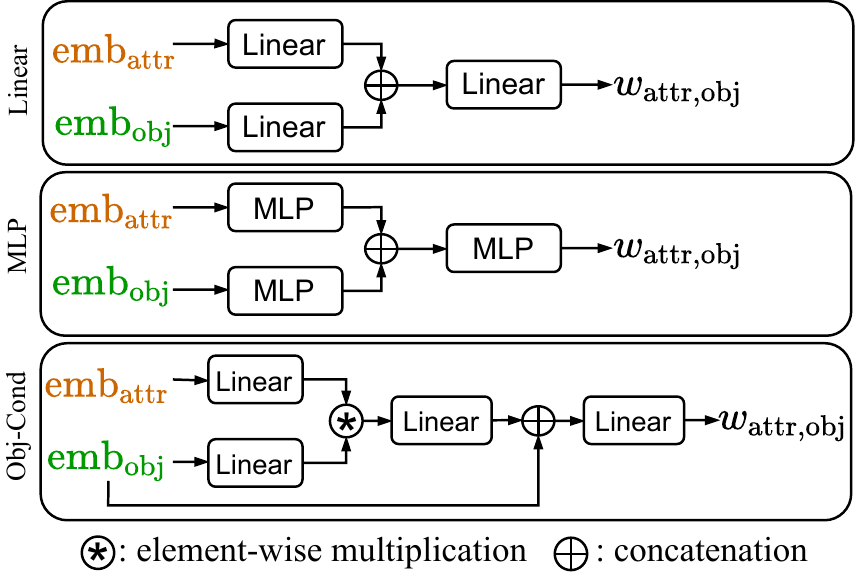}
\caption{We show the different networks used on top of word embeddings. Empirically and following our intuition, Object-Conditioned network works best among the three (rest two are Linear and MLP). (Sec~\ref{abl1})} 
\label{fig:emb_net}
\end{figure}

\noindent We experiment with different networks on top of word embeddings, namely Linear, MLP and Object-Conditioned. Object conditioned network uses word embedding for object to concatenate with attribute-object composition embeddings. We show in Figure~\ref{fig:emb_net}, the diagrammatic representation of different networks.
\subsection{Values for $\lambda$ and $\delta$}
\label{abl2}
\noindent We find the temperature variables $\lambda$ and $\delta$ empirically. The values $\lambda = 10$ and $\delta = 0.05 $ works best for OADis. Table~\ref{tab:word_appendix} shows the results for all the different configurations. To understand the effect of each temperature variable, we keep all the rest of the parameters constant and only change the studied parameter.
\begin{figure*}
\centering
\includegraphics[width=0.95\linewidth]{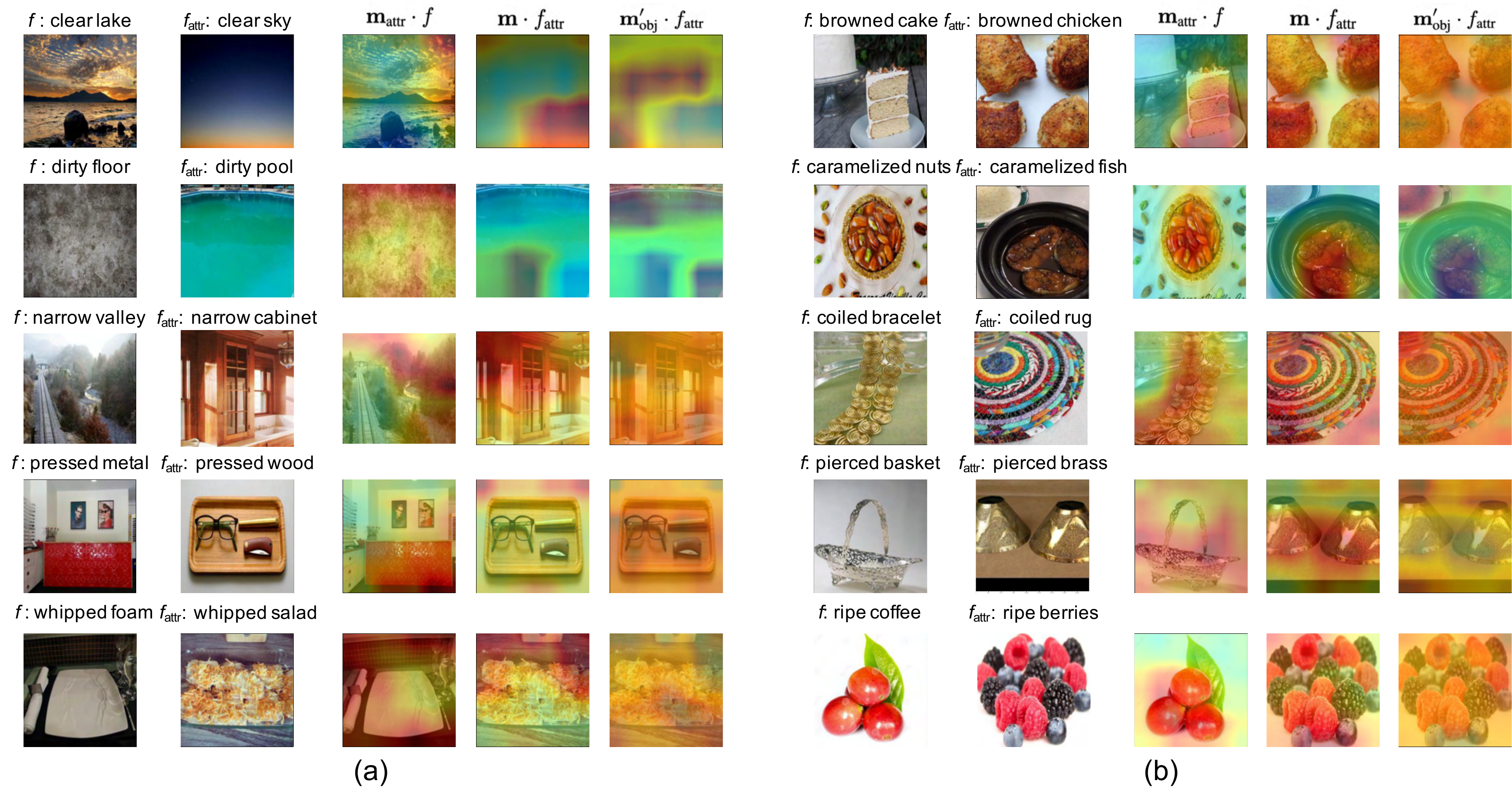}
\caption{\textbf{(a) Failure Cases:} Shows the image pairs, $f$ and $\fat$, and the similarity and dissimilarity map overlayed (details in Sec~\ref{at-mit}). Moreover, we show for some cases for MIT-States, the examples are very vague or incorrect to actually capture attribute and object concepts separately. For instance, in \texttt{clear lake} and \texttt{clear sky}, it is very difficult to distinguish lake and sky. Hence the similarity and dissimilarity maps do not perform very well. Other examples are also of failure cases where the overlayed similarity and dissimilarity maps do not make sense. \textbf{(b) Correct Examples:} This shows some good examples, where the similarity and dissimilarity maps capture the attibutes and objects correctly for MIT-States. } 
\label{at-mit}
\end{figure*}%
\begin{figure}
\centering
\includegraphics[width=0.95\linewidth]{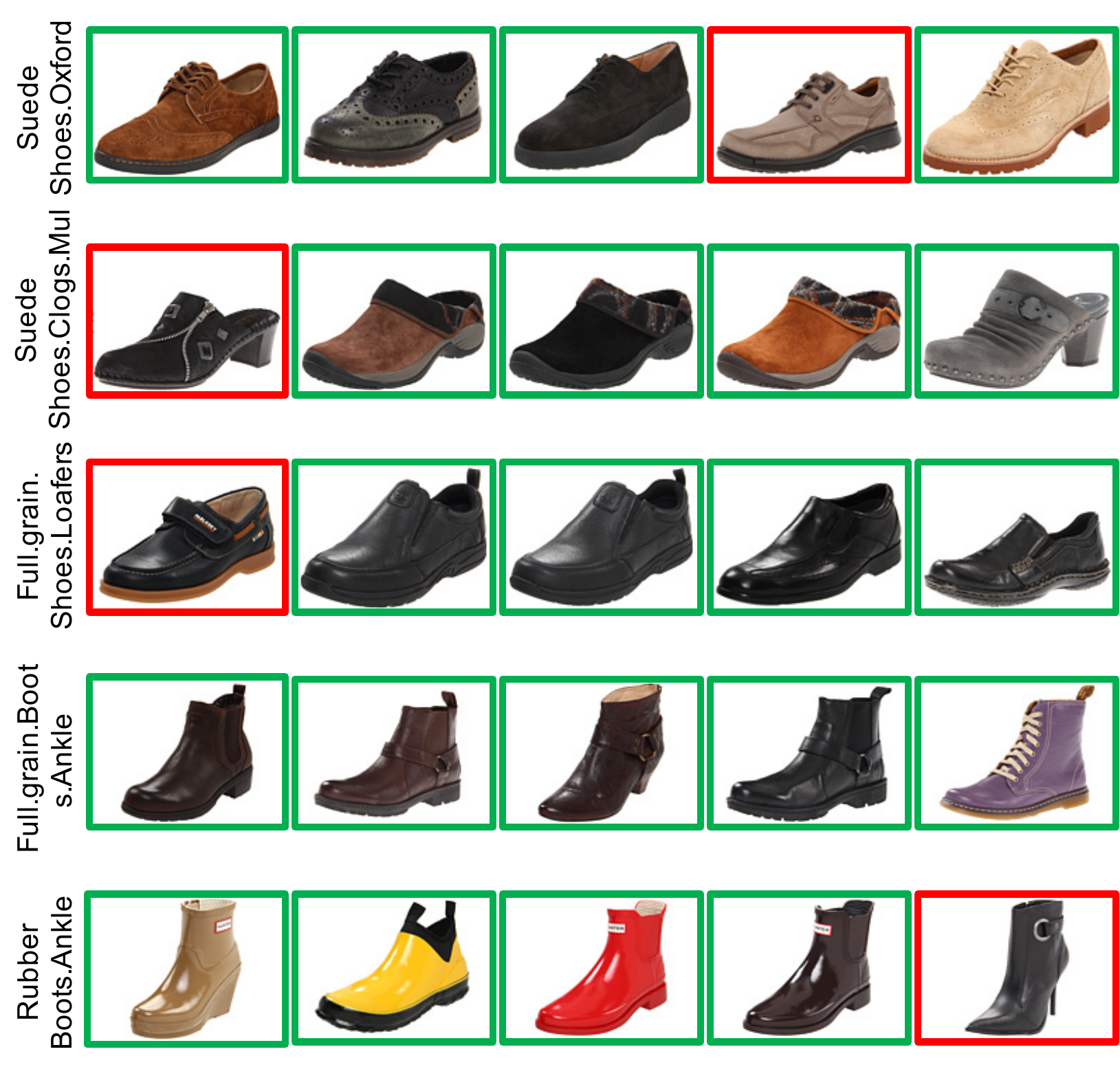}
\caption{We show the top 5 nearest neighbors using the hallucinated unseen composition features for
UT-Zappos. All the neighbors with correct labels are represented by \green{green}, whereas incorrect ones are represented with \red{red} outline.} 
\label{ut}
\end{figure}
\begin{table}
\caption{Results with pre-trained word-embeddings. GloVe~\cite{glove} performs the best, and is therefore used for OADis. (Sec~\ref{abl2})}
 \label{tab:word_appendix}
  \centering
  \small
  \vspace{-0.1in}
\resizebox{0.55\columnwidth}{!}{
 \begin{tabular}{@{}lcc@{}}
 \toprule
$\lambda$& Val AUC@1 & Test AUC@1 \\
\midrule
0.01 & 7.5 & 5.6 \\
0.1	&7.5&	5.7 \\
1&7.4	&5.7\\
\textbf{10}&	\textbf{7.6}&	\textbf{5.9}\\
100&	7.4	& 5.7\\
\midrule 
$\delta$& Val AUC@1 & Test AUC@1 \\
\midrule
0.01 & 6.4 & 4.8\\
\textbf{0.05} & \textbf{7.6} & \textbf{5.9}\\
0.1 & 6.7 & 5.2\\
\bottomrule
\end{tabular}}
\end{table}
\subsection{Different weights for losses}
\label{abl3}
\noindent We mention different weights for each loss function in the paper, in Section 3.3. Each $\alpha$ value is empirically found, and is used in the following equation for final loss function:
\begin{equation*}
\begin{split}
 \mathcal{L} &= \lcls + \alpha_1 \lat + \alpha_2 \lob + \alpha_3 \lse  + \alpha_4 \lun
 \end{split}
 \end{equation*}
Note that $\lcls$ is the main branch. The object and attribute losses are complementary, as shown in paper (Table 4). Hence, $\alpha_1$ and $\alpha_2$, which are the weights for $\lat$ and $\lob$ share the same values, \ie 0.5. Finally, $\alpha_4$ and $\alpha_5$ have the same value since both are composition losses for seen and unseen pairs, \ie 0.05. The chosen weights for $\alpha$ values are in bold in Table~\ref{tab:al}.

\begin{table}
\caption{We show empirical weights of each loss function in this table. (Sec~\ref{abl3})}
 \label{tab:al}
  \centering
  \small
  \vspace{-0.1in}
  \resizebox{\columnwidth}{!}{
 \begin{tabular}{@{}ccccc@{}}
  \toprule
$\alpha_1$  and $\alpha_2$ & $\alpha_3 $ and $\alpha_4$ & Val AUC@1 & Test AUC@1 \\
\midrule
0.1	& 0.05	& 7.1 &	5.7 \\
0.5 & 0.1 & 7.0 & 5.3 \\
0.1 &	0.05 &	7.5 & 5.8 \\
\textbf{0.5} & \textbf{0.05} &	\textbf{7.6} & \textbf{5.9}\\
1.0 & 0.05 & 7.3 &	5.6\\
\bottomrule
 \end{tabular}}
 \end{table}
\section{Qualitative results}
\noindent We show more qualitative results to support our architecture for different datasets.
\subsection{UT-Zappos.}
\noindent We show nearest neighbor results in paper for MIT-States~\cite{mitstates} (Fig. 4(a)). Here, we show similar study for UT-Zappos~\cite{utzappos} in Figure~\ref{ut}. Using the hallucinated composed features of unseen pairs, we find the top 5 nearest neighbors from test set. The red boxes show incorrect labels, where green show the correct labels. 
\subsection{Attention Maps}
\begin{figure}
\centering
\includegraphics[width=0.95\linewidth]{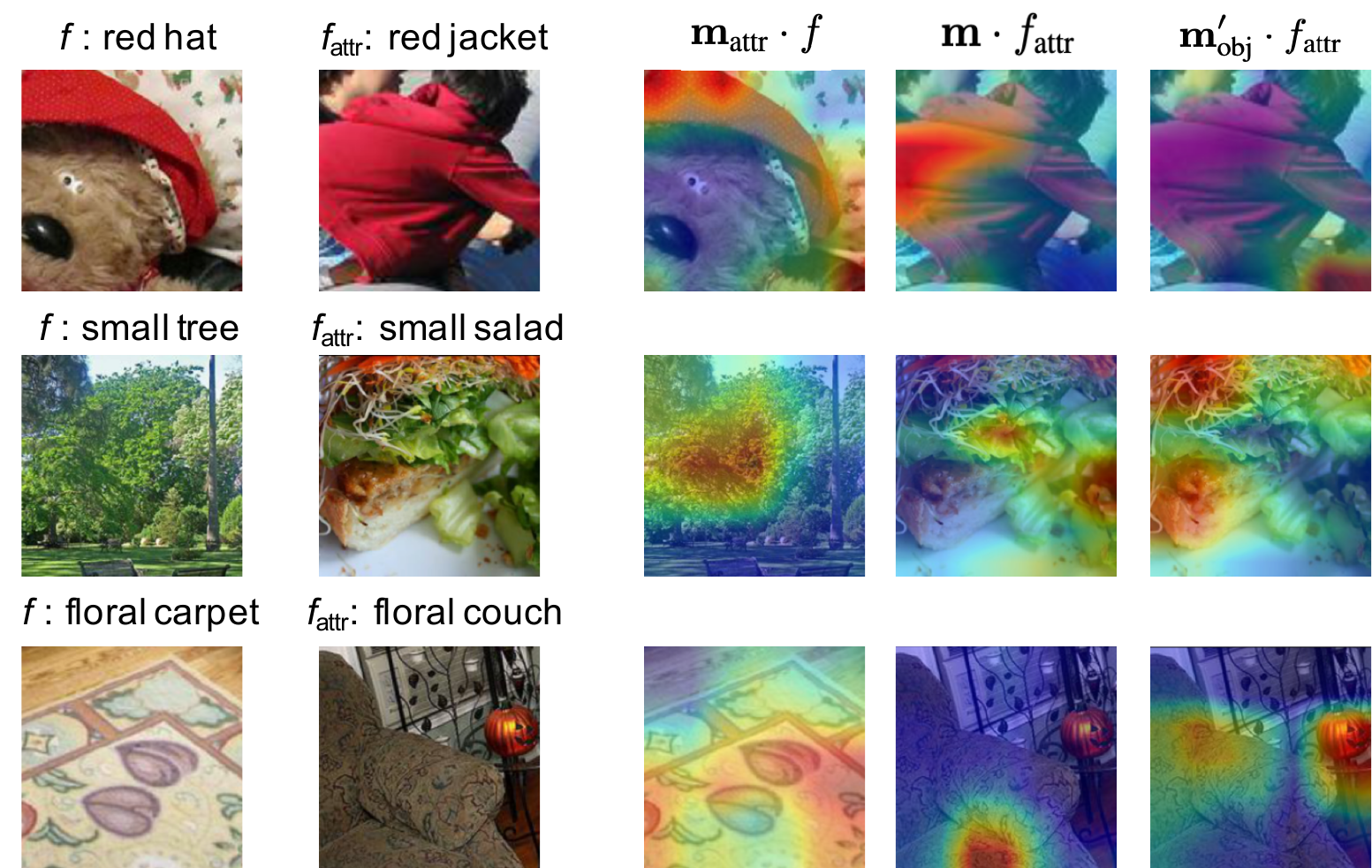}
\caption{\textbf{Correct Examples:} We show the similarity and dissimilarity attention maps overlayed on images for VAW-CZSL as well. To re-iterate, for images with features $f$ and $\fat$,  $\mathbf{\ma} \cdot f$ shows how the regions in $f$ which are most similar to $\fat$, and $\mathbf{m} \cdot \fat$ shows the regions in $\fat$ which are most similar to regions in $f$. Lastly, $\mathbf{\mop} \cdot \fat$ shows the regions of $\fat$ which are most dissimilar to $f$.} 
\label{vaw}
\end{figure}

\noindent In Figure~\ref{at-mit} and~\ref{vaw}, we show the qualitative results on MIT-States~\cite{mitstates} and VAW-CZSL, with examples $f$ and $\fat$ and overlayed feature maps. To re-iterate, for images with features $f$ and $\fat$,  $\mathbf{\ma} \cdot f$ shows how the regions in $f$ which are most similar to $\fat$, and $\mathbf{m} \cdot \fat$ shows the regions in $\fat$ which are most similar to regions in $f$. Lastly, $\mathbf{\mop} \cdot \fat$ shows the regions of $\fat$ which are most dissimilar to $f$. Although, the overlayed attention maps for similarity and dissimilarity make sense most of the times ( Figure~\ref{at-mit}(b)), due to some inconsistencies in dataset, we still find some samples where is it difficult to disentangle the attribute and object features.
The main reasons why this happens is:
\begin{itemize}
\setlength{\itemsep}{1pt}
  \setlength{\parskip}{0pt}
  \setlength{\parsep}{0pt}
  \item Some concepts are abstract, such as \texttt{clear sky, pressed metal, dirty floor} (fig.~\ref{at-mit}(a)), since it is very difficult to separate \texttt{dirty} from \texttt{floor}. Hence, the attention maps for similarity and dissimilarity do not make much sense.
  \item Some images in MIT-States and even in other dataset are mislabelled (\eg \texttt{whipped foam} in fig.~\ref{at-mit}(a)), which makes it  difficult to learn attributes from those.
  \item Finally, for some cases, like \texttt{narrow valley}, our method fails to disentangle attribute and object similarity, due to various objects in the scene. For future work, using a foreground and background separator before finding similarities and dissimilarities between features can be helpful.
\end{itemize}
  
\vspace{-1.0em}
\section{Negative Impact of our work}
\noindent Our work is a new initiative in the direction of learning visual features for objects and it's attributes. We present it as a prototype, or an alternative direction for understanding attributes-object pairs. Similar to any other work in vision, learning attributes of objects can have various positive implications, \eg in object detection, knowing attributes can provide additional knowledge about the objects. However, knowing the additional information about attributes, it can be used for persuasion for marketing policies, for even worse factors. Even though it seems very far fetched ideas, but using attribute classification along with object detection, knowing the attributes people can build weapons and ammunition to either counter attack the present ammunition. Attribute classification can also be used on humans, to detect certain traits of human for bypassing large-scale surveillance applications. In general, attributes provide additional information for objects, which can be used negatively or positively. 

\section{Dataset license}
Because we are creating the VAW-CZSL dataset based on the existing VAW dataset, as per the guideline of CVPR 2022, we provide the VAW dataset URL and license as follows:
\begin{itemize}
    \item URL: \url{https://vawdataset.com}
    \item License: \url{https://github.com/adobe-research/vaw_dataset/blob/main/LICENSE.md}
\end{itemize}

\end{document}